\documentclass{article}
\usepackage{fullpage}

\usepackage{cite}
\usepackage{comment}
\usepackage{amsmath,amssymb,amsfonts}
\usepackage{algorithmic}
\usepackage{graphicx}
\graphicspath{{figures/}} 
\usepackage[export]{adjustbox} 
\usepackage{textcomp}
\usepackage{xcolor}
\usepackage{booktabs}
\usepackage{multirow}
\def\BibTeX{{\rm B\kern-.05em{\sc i\kern-.025em b}\kern-.08em
    T\kern-.1667em\lower.7ex\hbox{E}\kern-.125emX}}

\usepackage{tikz}

\newcommand{\word}[2]{w_{#1, #2}} 
\newcommand{\wordrep}[2]{v_{#1, #2}} 
\usepackage{hyperref}

\begin{document}

\title{\bf  Learning language variations in news corpora \\ through differential embeddings}

\date{}

\author{Carlos Selmo$^{1}$ \and Julian F. Martinez$^{2,3}$ \and Mariano G. Beir\'o$^{2,4}$ \and and J. Ignacio Alvarez-Hamelin$^{2,4}$\\
\small{$^{1}$Instituto Tecnol\'ogico de Buenos Aires, Buenos Aires, Argentina. {\tt cselmo@itba.edu.ar}} \\
\small{$^{2}$Universidad de Buenos Aires, Facultad de Ingenier\'{\i}a. Buenos Aires, Argentina.}\\
\small{$^{3}$CONICET -- Universidad de Buenos Aires. Instituto de C\'alculo. Buenos Aires, Argentina.}\\
\small{$^{4}$CONICET -- Universidad de Buenos Aires. INTECIN. Buenos Aires, Argentina.}\\
\small{\tt cselmo@itba.edu.ar  \{jfmartinez,mbeiro,ihameli\}@fi.uba.ar}
}

\maketitle

\begin{abstract}
There is an increasing interest in the NLP community in capturing variations in the usage of language, either through time (i.e., semantic drift), across regions (as dialects or variants) or in different social contexts (i.e., professional or media technolects).
Several successful dynamical embeddings have been proposed that can track semantic change through time. 
Here we show that a model with a central word representation and a slice-dependent contribution can learn word embeddings from different corpora simultaneously.
This model is based on a star-like representation of the slices.
We apply it to The New York Times and The Guardian newspapers, and we show that it can capture both temporal dynamics in the yearly slices of each corpus, and language variations between US and UK English in a curated multi-source corpus.
We provide an extensive evaluation of this methodology. 
\end{abstract}


\section{Introduction}

Word embeddings like Word2Vec~\cite{SGNS, mikolov2013linguistic} or Glove~\cite{pennington2014glove} can learn context-sensitive vector representations of words from very large corpora. These representations have proven useful for supervised tasks like language translation, entity recognition, sentiment analysis, or question answering.

The more general problem of tracking the semantic change of words through time has initially been addressed by a number of works, either by connecting several static embeddings through mapping transformations, or by initializing the training of each slice with the results from the previous one in the Word2Vec case (e.g.,\cite{kim-etal-2014-temporal, kulkarni2015statistically, Hamilton2016}).
More recent works can deal with all the temporal slices simultaneously, as in Bamler and Mandt~\cite{bamler2017dynamic}, Rudolph and Blei~\cite{rudolph2018dynamic}, and Yao et. al~\cite{yao2018dynamic}.  These works link the slices either by a diffusion process, a random walk, or a regularization term in the cost function.

The proposed approach does not assume any sequentiality in the slices (as in diachronic embeddings). 
By combining data from different sources, these embeddings can help to understand not only semantic drift, but also cross-cultural differences (e.g., British vs. American English) or dialect variations (e.g., regional dialects in Twitter~\cite{grieve2019mapping}).

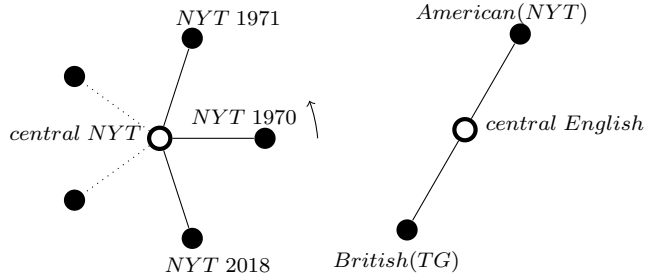
\begin{figure}[t]
    \centering
    {\footnotesize
    \begin{tikzpicture}[scale=0.7]
        \node[draw, circle, ultra thick] at (360:0mm) (center) {};
        \node at (-1.6,0.1) {$central\;NYT$};

        \draw[->]     (0:3.0cm) arc (5:20:2.8cm);

        \node[circle,fill=black] at ({1*360/5}:2cm) (11) {};
        \draw (center)--(11);
        \node at (1.3,2.3) {$NYT\;1971$};

        \node[circle,fill=black] at ({2*360/5}:2cm) (22) {};
        \draw[dotted] (center)--(22);

        \node[circle,fill=black] at ({3*360/5}:2cm) (33) {};
        \draw[dotted] (center)--(33);

        \node[circle,fill=black] at ({4*360/5}:2cm) (44) {};
        \draw (center)--(44);
        \node at (1.1,-2.4) {$NYT\;2018$};

        \node[circle,fill=black] at ({5*360/5}:2cm) (55) {};
        \draw (center)--(55);
        \node at (1.6,0.4) {$NYT\;1970$};

    \end{tikzpicture} \begin{tikzpicture}[scale=0.7]
            \node[draw, circle, ultra thick] at (360:0mm) (center) {};
            \node at (1.9,0.1) {$central\;English$};

            \node[circle,fill=black] at ({1*360/6}:2.1cm) (111) {};
            \draw (center)--(111);
            \node at (0.7,2.2) {$American (NYT)$};

            \node[circle,fill=black] at ({4*360/6}:2.2cm) (133) {};
            \draw (center)--(133);
            \node at (-1.3,-2.5) {$British (TG)$};

        \end{tikzpicture}
}
\caption{Star-like representation for \textit{(left) temporal embeddings in The New York Times (NYT) Corpus}; \textit{(right)} cross-dialectal embeddings for the NYT, American English, and The Guardian (TG), British English.}\label{fig:star}
\end{figure}

In this work we consider that each corpus is divided into a set of segments called \textit{slices}. All the slices are trained simultaneously, following the Word2Vec distributional hypothesis, with the addition that each word vector representation inside a slice is obtained by adding a central representation and a slice-dependent one. Thus, the different representations of one same word across different slices are tied by a common component. This constraint can be depicted as a star-like graph.
Figure~\ref{fig:star} shows this representation for two cases: (left) a newspaper corpus through covering several years, and (right) a multi-source corpus combining two English-language newspapers.

The rest of the paper is organized as follows. 
Section~\ref{sec:new_model} introduces the proposed model, giving its formal description, vocabulary selection and implementation details.
Section~\ref{sec:data} describes the datasets used for this work: two corpora from The New York Times and The Guardian newspapers, and a curated multi-source corpus that combines both of them.  
Section~\ref{sec:exp} provides experimental work on the three datasets, and their corresponding quantitative and qualitative analysis.
The related work is detailed in Section~\ref{sec:prev}. 
Finally, our conclusions and future work are discussed in Section~\ref{sec:conc}.

\section{Previous works}\label{sec:prev}

We start by describing those works which are close to our proposal, either in the type of situations they are dealing with or in its methodology. 
As far as we are aware, the first work in dealing with contextual information is \cite{bamman2014distributed}. 
While formally our proposal follows nearly the one presented by the authors, from a methodological point of view it differs in many aspects.  
In \cite{bamman2014distributed} the method is tested only on US geolocated (by state) tweets, representing states as contexts (herein, slices).
For the training the authors use hierarchical softmax. 
Finally, succinct qualitative and quantitative evaluations are done, by inspecting the neighborhood of some selected words.\\
In \cite{kulkarni2016freshman}, the same model is considered, but with the distinction that the use of a null model is included to decide (via a particular score) whether a change in the representation of a word is statistically significant. 

We describe now the previous work within Dynamic Word Embeddings.  
Firstly, we detail works in which the proposed methods therein \emph{train each time slice separately}.

To the best of our knowledge, the first work in dealing with Dynamic Word Embeddings is  \cite{kim-etal-2014-temporal}.
The authors train sequentially a skip-gram model (word2vec ~\cite{mikolov2013linguistic}) for each year slice; where each year embedding serves as an initialization for the training of the next year embedding.  
They use the method for automatic detection changes in language on the Google Books N-gram corpus.
Similarly, the method proposed in \cite{Kulkarni2015} follows closely the previous one, with the distinction of an \emph{alignment process} between subsequent years embedding by learning a linear transformation via a piece-wise linear regression model. 
They do also focus in linguistic changes by analyzing the Google Books N-gram corpus and movie reviews from Amazon. 
Finally, in \cite{Hamilton2016} they make use of three different methods to construct the embeddings within each time-period: PPMI, SVD, and SGNS (all of them related to word2vec) \cite{levy2015improving, kim-etal-2014-temporal}.
The alignment between subsequent time-period representations is done through orthogonal Procrustes. 
They explore two statistical laws (conformity and innovation) relating frequency and polysemy to semantic change for the Google Books N-gram and COHA corpora. 

We now switch to works in which the proposed methods therein \emph{train across several time periods jointly}. 

In \cite{bamler2017dynamic}, the embedding vectors are inferred from a probabilistic version of word2vec. 
These embedding vectors are connected in time through a latent diffusion process.
This algorithm was ran on Google Books, State-of-union (addresses of U.S. presidents), and a news tweets set extracted from Twitter. 
They performed quantitative and qualitative analysis (taking \cite{kim-etal-2014-temporal} and \cite{Hamilton2016} as baselines), finding improvements regarding the predictive likelihoods on held-out data while smoother embedding trajectories.

Rudolph and Blei \cite{rudolph2018dynamic} presented a dynamic embedding based on an extension of a Bernoulli embedding.
Smoothness along time slices is obtained by using a Gaussian random walk as a prior on the embedding vectors.  
Their method is evaluated on ArXiv-Machine Learning articles, ACM-abstracts, and U.S. Senate-Speeches.

In \cite{yao2018dynamic}, based on the relation established in \cite{levy2014neural}, the authors propose a method reminiscent of the PPMI factorization by solving an optimization problem while enforcing the embeddings alignment via a regularization term which penalizes the differences between two slice subsequent factorizations. 
Their method is tested on the New York Times corpus (from 1990 to 2016), with the purpose of working with a corpus that maintains consistency in narrative style and grammar. 
New qualitative and quantitative evaluations are proposed. 
The qualitative ones are: visualization of the trajectory of a word in the embedded space, word's equivalence searching, and word's popularity determination.
The quantitative ones look over semantic similarity, alignment quality, and robustness. 
Some of these criteria have been used to evaluate our model.

Barranco et al.~\cite{barranco2018tracking} present an embedding based on transformed tf-idf weights.
The transformation makes use of a Gaussian filter to diffuse the contribution of each document to the regular tf-idf weights, before and after its publication date.
They introduce the concept of neighborhood monotony to evaluate, quantitatively, how much the context of a word changes over time.
They compare their model with a regular tf-idf model and the one proposed in \cite{rudolph2018dynamic} by analyzing the PubMed abstracts corpus.

Finally, the work of Kutuzov et al.~\cite{kutuzov-etal-2018-diachronic} presents an extensive analysis of the different dynamic word embeddings, focusing specially in their use for semantic shifts detection.

\section{A multi-source word embedding}\label{sec:new_model}

We propose a \textit{multi-source word embedding} method (MW2V) which produces \emph{multiple} linked word vector representations  from different text sources (\emph{slices}), endowed with the following property:
\begin{description}
\item 
If the word does not reveal semantic change between slices, its representations on those slices should be nearly identical, while the representations among slices, where the word does exhibit semantic change, must present perceptible deviations.
\end{description}
The presumption behind this property is that most of the words do not have major semantic differences throughout the slices. 
This idea applies for a variety of contexts, where slices may represent 
time, dialects, or field-specific knowledge.

We present our model using a set-up and notation that follows closely those on \cite{SGNS, Hamilton2016}. 

The corpus for slice $s$ is denoted as a sequence of words 
$\word{s}{1}, \word{s}{2},$ $\ldots, \word{s}{T_s}, \quad \word{s}{i} \in V_s$ (the vocabulary for slice $s$).
We model the distribution of each sequence by the following conditional model, in which the probability of appearance of an \emph{output} (\emph{context}) word $w_s^O$ given its \emph{input} (\emph{target}) word $w_s^I$ is given by
$$
p(w_s^O|w_s^I) 
=
\frac{
\exp{(
\wordrep{s}{w_s^O}'^{\ \mathrm{T}} \ \wordrep{s}{w_s^I}
)}
}{\sum_{\Tilde{w} \in V_s}
\exp{(
\wordrep{s}{\Tilde{w}}'^{\ \mathrm{T}} \ \wordrep{s}{w_s^I}
)}
}\enspace,
$$
where $\wordrep{s}{w}, \  \wordrep{s}{w}' \in \mathbb{R}^d$ are the vector representations of word $w$ in slice $s$ as context/target respectively 
(with $d\ll |V|, \quad V :=\cup_s V_s$) and $\mathrm{T}$ denotes the transpose of a vector.

\subsection{Method description}

From the aforementioned presumption, we propose the following word vector representation:
\begin{equation*}
v_{s,w} = \mathring{v}_{w} + \delta_{s,w}\enspace,
\qquad 
v_{s,w}' = \mathring{v}_{w}' + \delta_{s,w}'\enspace,
\end{equation*}
where:
$\mathring{v}_{w}$ (resp. $\mathring{v}_{w}'$) is the common vector of a word across all the slices and 
$\delta_{s,w}$ (resp. $\delta_{s,w}'$) is the drift from the common vector for each particular slice $s$ (see Figure~\ref{fig:star}). 

To obtain the parameters of this model, we maximize a regularized and tractable version of the \emph{pseudo-log-likelihood}:
$$
\mathcal{L}(\{\mathring{v}_{w}, \delta_{s,w}; \ \mathring{v}_{w}', \delta_{s,w}'\}_{s,w}) =
\sum_{s=1}^S [
\mathcal{L}^s_{pos} + \mathcal{L}^s_{neg} + \mathcal{L}^s_{reg}]\enspace, 
$$
with
\begin{align}
\mathcal{L}^s_{pos} & =     
\frac{1}{T_s} \sum_{t=1}^{T_s}\sum_{-c \leq j \leq c, j \neq 0} 
\log \sigma \big(\wordrep{s}{\word{s}{t+j}}'^{\ \mathrm{T}} \wordrep{s}{\word{s}{t}}\big) 
\enspace, \label{eq:pos}\\
\mathcal{L}^s_{neg} & = 
\frac{1}{T_s} \sum_{t=1}^{T_s}
\sum_{k=1}^K \mathbb{E}_{P_n(\cdot|s)} 
\left[  
\log \sigma\big( -\wordrep{s}{W^k_O}'^{\ \mathrm{T}} \wordrep{s}{\word{s}{t}}\big)
\right]\enspace, \label{eq:neg} \\
\mathcal{L}^s_{reg} & = 
-\sum_{w \in V_s} \lambda_{s,w}   \|\delta_{s,w}\|^2 \enspace, \label{eq:reg}
\end{align}
where  $\sigma(z):=\tfrac{1}{1+e^{-z}}$, $c$ is the size of the training context, $W^k_O$ are independent random words sampled from $P_n(\cdot | s)$, the distribution obtained by 
\begin{equation}
P_n(w|s) \propto U_s(w)^{\frac{3}{4}} \enspace, \label{eq:n_s}
\end{equation}
with $U_s(w)$ the unigram distribution of the slice $s$ and $\|\cdot\|$ is the euclidean norm of a vector in $\mathbb{R}^d$.
The abbreviations $pos, neg$ refer to positive and negative samples, while $reg$ refers to a regularization term.

For the case of temporal slices, we can adjust $\lambda_{s,w}$ according to~\cite{Hamilton2016}:
\begin{itemize}
\item {\bf The law of conformity}: 
frequent words change more slowly,
\item {\bf The law of innovation}: 
polysemous words change more quickly.
\end{itemize}

For simplicity along this work, we will assume $\lambda_{s,w}= \lambda$ as a constant hyperparameter.

\subsection{Implementation}
 
\begin{figure}[th]
    \centering
    \includegraphics[trim=0 150 0 180,clip,width=0.5\columnwidth]{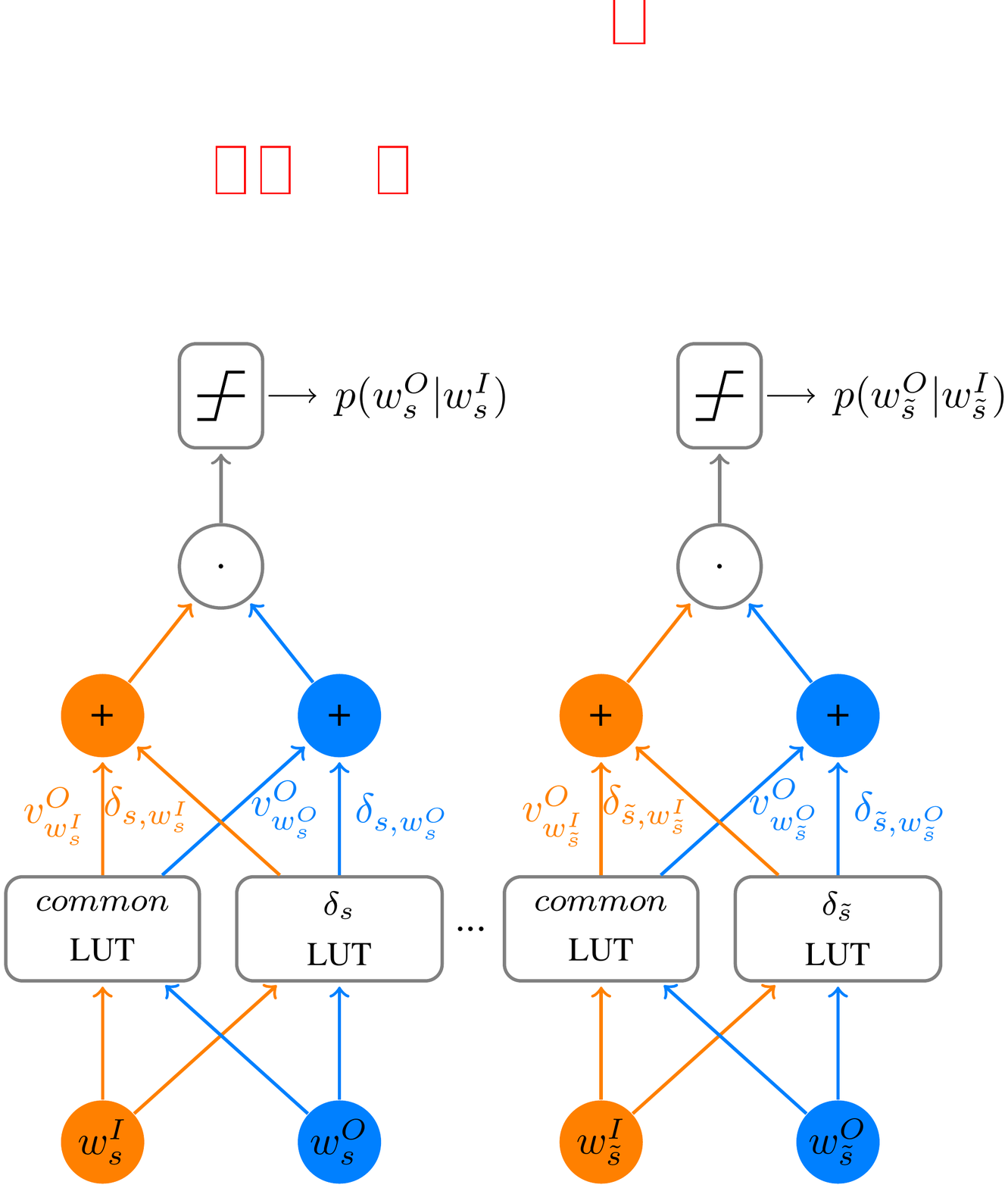}
    \caption{Neural Network architecture for the MW2V. Each vertical block represents a slice, and each word pair (target $w^I$ and context $w^O$), feeds the corresponding slice from which it is sampled. For simplicity we represented only two slices, $s$ and $\tilde{s}$. The common $LUT$ block is the same one for every slice.} 
    \label{fig:arch}
\end{figure}

We implemented the model as a neural network whose architecture is sketched in Figure~\ref{fig:arch}.
Each pair of context $w^O_s$ and target $w^I_s$ words from slide $s$ are feed to the lookup table ($LUT \in \mathbb{R}^{d \times |V|}$) neural network, optimized according to Equations~\ref{eq:pos},~\ref{eq:neg}, and~\ref{eq:reg}. 
The common $LUT$ gives the central representation for each word, while the $\delta LUT_s$ gives the difference component for each word for the slice $s$.

In this way, we optimized the training time since each slice was trained in parallel, taking advantage of the GPU parallelization. 
The model was implemented using Keras with TensorFlow backend~\cite{chollet2015}. 

\bigskip

For each slice we build a vocabulary $V_s$ from the most frequent words in its unigram distribution. 
From the union of all these vocabularies, we define a global vocabulary: $V$.  
Thus there are words which are present in the aforementioned $LUT$'s while not in the corpus of the slice $s$.

The proposed common $LUT$ implies a global indexing of the vocabulary. 
On the other hand, a usual way to perform the word sampling efficiently requires indexing the vocabulary according to its unigrams distribution ranking. 
Thus, the existence of several vocabularies calls for a convenient adaptation.
Due to this observation we proceed as follows:
\begin{itemize}
\item 
We define one indexing for each vocabulary, according to its unigrams distribution, a \emph{slice indexing} $g_s: V_s \rightarrow \{1,\ldots, |V_s|\}$.
Additionally, we consider a \emph{global indexing} $g : V \rightarrow \{1,\ldots, |V|\}$ and with this two indexing, we obtain a mapping between local indices and global ones by $idx_s(\cdot):= g(g_s^{-1}(\cdot))$.
\item 
We start $\delta LUT_s[r, g(w)]=0$, for all $w \notin V_s$, $r=1, \ldots, d$. This condition remains unchanged during training because of our customized sampling scheme per slice. 
Therefore the common representation of each word does not get modified by those slices where the word is not present. 
\end{itemize}

Each epoch consists of sub-samples of positive pairs of words, taken with a sampling factor of $10^{-5}$ as in \cite{mikolov2013linguistic}.
For each four positive samples, three negative samples are taken according to Equation~\ref{eq:n_s}.

To train the model we used the Adam optimizer~\cite{kingma2014adam} with Cyclical Learning Rate (CLR,~\cite{smith2017cyclical}), a context size $c=4$ and a fixed value of $\lambda=10^{-9}$ for all the slices (Equations 1, 2, y 3).
The use of CLR reduced training time.

Let us remark that with these adaptations we obtain that whenever $\lambda>>1$, we get  $|\delta_{s,w}| \simeq 0$ for all $w \in V, s$, while when $\lambda =0$ we obtain the same representations as training, independently, one word2vec per slice.

\section{Datasets}\label{sec:data}

Our proposition deals with slices; then, we provided various datasets to test two possibilities: slices as time and slices as different versions of English language. 
We selected three datasets to test the MW2V. 
The first two are The New York Times (NYT) and The Guardian (TG) newspapers, where we used a one-year per slice representation to capture semantic evolution in arguments like politics, external affairs, or oil.
The third dataset is a combination of both newspapers, aimed at capturing cultural differences in the English language.

\paragraph{The New York Times (NYT)}
We obtained the NYT dataset provided by its public API\footnote{https://developer.nytimes.com} filtered by the following \texttt{type-of-material}: News, Article, Editorial, Letter, or Blog.
We grouped articles from 1990 to 2016 into one-year slices, with $65,972$ articles per year on average. 
To train the model, we chose a vocabulary size of $|V_s|=20,000$ words per slice, obtaining a total vocabulary $V=\bigcup_{s}{V_s}$ with $39,218$ terms. 
Our embedding contains $d=48$ dimensions. 
It took 6 hours to train the model using an NVidia Titan XP GPU.

\paragraph{The Guardian (TG)}
We got the TG corpus from the public API\footnote{https://open-platform.theguardian.com}, and trained our model using one-year slices from 1999\footnote{1999 is the first available year in TG.} to 2016, selecting articles from the following \texttt{type-of-material}: Opinion, World News, Sports, Football, Art, Business, Money, Life and Style, Fashion, Books, Film, Technology, Stage, Food, Science, Travel or Education.
We chose a vocabulary size of $|V_s|=20,000$ words per slice. 
The total vocabulary size is $|V|=32,371$ and the embedding has $d=48$ dimensions. 
As same as the NYT, its model training needed 6 hours in the same GPU.

\paragraph{NYT-TG}
We combined the two newspapers to highlight the use of different sources of the English language. 
Here slices do not represent time but a geographic source. 
We selected the articles between 2010 and 2016 and built two slices: one is the NYT, and the other is TG. 
We used a vocabulary size of $|V_s|=10,000$ words per slice while the total size is $|V|=16,528$. 
This embedding has $d=32$ dimensions, and its training took 1 hour in the same GPU.

\section{Experimental analysis}\label{sec:exp}

The evaluation of word embeddings is not a standardized task, and different methods have been suggested for it.
In this section we propose to use a set of metrics to evaluate the semantic quality from a quantitative perspective, and some others methods for analyzing the qualitative behavior of our proposed MW2V.

\subsection{Quantitative Evaluation}

The quantitative analysis is based on Yao et al.~\cite{yao2018dynamic}, who defined a set of metrics to test some characteristics of embeddings, and applied them to NYT corpus. 

The first type of analysis is the \emph{semantic similarity}; that is, how well an embedding propagates the meaning of a word throughout slices. 
For this purpose, the authors propose a partition of the vocabulary based on the NYT sections, and check how different it is to the one obtained by clustering the embedding representations.

More precisely, for each slice $s$ and word $w \in V_s$ they assigned a section $l_{s,w} \in \{1,\ldots,L\}$, $L$ being a numbering of the pre-selected sections.
The set of \emph{triplets}, $\{(s,w,l_{s,w})\}_{s,w}$ define a partition $\mathcal{C}^0:= \bigcup_{l=1}^L \mathcal{C}^0_l$, where $\mathcal{C}^0_l:=\{(s,w) \ : \ l_{s,w}=l \}$.

For each dataset used, the sections used in aforementioned triplets were chosen as follows:
\begin{itemize}
	\item{\bf NYT:} Business, Sports, Arts, U.S., World, Fashion and Style, Technology, Health, Science, Real Estate, Home and Garden.
	\item{\bf TG:} Opinion, World News, Sports, Football, Art, Business, Money, Life and Style, Fashion, Books, Film, Technology, Stage, Food, Science, Travel, Education.
	\item{\bf NYT-TG:} (e.g., NYT\_section--TG\_section) Opinion--Opinion, Word--World\_News, Sports--(Sports,Football), Arts--Arts,
	
	Business\_Day--(Business,Money), Fashion\_and\_Style-
	
	-(Life\_and\_Style,Fashion), Books--Books, Movies--Films, 
	
	Technology--Technology, Theater--Stage, Food--Food, Science--Science, Travel--Travel, Education--Education. 
\end{itemize}
Notice that in the last case, some sections of NYT corresponds to several of TG (indicated between brackets). 
For each combination of section and year, we picked those words among the 200 most popular such that their relative frequency within this section for this year is greater than $35\%$ (see~\cite{yao2018dynamic}).
The number of triplets
for the NYT, the TG and the NYT-TG datasets are 1526, 2191 and 2177, respectively. 

On the other hand, they obtained another partition $\mathcal{C}^*$ by applying $k$-means algorithm (with cosine similarity) for the embedding representations $\{v_{s,w}\}_{s,w}$.
Notice that this information is not used during training, therefore this approach is a way to measure the quality of the embedding.

We use the following metrics to evaluate the different embeddings, according to the work~\cite{yao2018dynamic}:
\begin{itemize}
\item Normalized  Mutual Information (NMI), 
defined by 
$$NMI(\mathcal{C}^0,\mathcal{C}^*)=\frac{2\cdot I(\mathcal{C}^0,\mathcal{C}^*)}{(H(\mathcal{C}^0)+H(\mathcal{C}^*))},$$ where $H(\cdot)$ is the entropy and $I(\cdot)$ is the mutual information.
\item The $F_\beta$-score, 
defined as $F_\beta=\frac{(\beta^2+1)\cdot P \cdot R}{\beta^2\cdot P+ R}$, with the precision $P=Tp/(Tp+Fp)$ and the recall $R=Tp/(Tp+Fn)$, where $Tp$, $Fp$, $Fn$ stands for "True-positive," "False-positive," and "False-negative." 
\end{itemize}  
Any pair of words $(s,w), (s',w')$ within the same cluster of $\mathcal{C}^*$ and such as $l_{s,w}= l_{s',w'}=l$ is counted as True-positive. 
Every pair of words which share the same section while belonging to a different cluster, is counted as a False-negative. 
Finally, each pair of words both of which are in the same cluster, but its sections are different is counted as a False-positive. 

For the $F_\beta$-score, we use the same value $\beta=5$ as in ~\cite{yao2018dynamic}, which penalizes the false-negative and gives more weight to the recall.

The second aspect is related to \emph{alignment quality}, that is, if the semantic distribution is consistent along with the slices. 
For example, it is expected that ``president'' does not change along time, but ``Bush'' or ``Obama'' move accordingly time changes.
In order to assess this behavior, an \emph{alignment test} is built for each data-set.
More precisely, an alignment test is a set of relations $\{(s,w) \stackrel{j}{\leftrightarrow} (s',w')\}_{j=1, \ldots, J}$.
For the NYT, we use the Yao et al.~\cite{yao2018dynamic} ones, while for the TG, we produce a list gathering some massive knowledge which presents some stability of the subject over time, e.g., political well known names of the United Kingdom\footnote{Similarities files can be obtained from \url{https://anonymous.4open.science/r/bac3e94b-6014-41d1-ba86-6f9c41171391/}}.
For the cross newspapers dataset (NTY-TG case), building a large enough and meaningful list of equivalences is far beyond the scope of this work.

For each $v \in \mathbb{R}^d$, we consider $N_k(v;s')$ the set of the $k$-closest embedding representations from the slide $s'$ (in the sense of cosine similarity).

Once again, we follow ~\cite{yao2018dynamic} by using the following metrics:
\begin{itemize}
\item Mean Precision@$k$, 
$$\bar{\mathcal{P}}_k=\frac{1}{J}\sum_{(s,w) \stackrel{j}{\leftrightarrow} (s',w')}{\mathcal{P}_k[j]},$$ 
where $\mathcal{P}_k[j]$ is $1$ if $v_{s',w'} \in N_k(v_{s,w}; s')$ and 0 otherwise.

\item Mean Reciprocal Rank, 
$$MRR=\frac{1}{J}\sum_{(s,w) \stackrel{j}{\leftrightarrow} (s',w')}{1/rank(j)},$$ 
where $rank(j)$ is the ranking of the representation $v_{s',w'} \in N_{10}(v_{s,w}; s')$ and $\infty$ if $v_{s',w'} \notin N_{10}(v_{s,w}; s')$.
\end{itemize}

We repeat the benchmarks proposed in Yao et al.~\cite{yao2018dynamic} for the proposed methodology and compare them  with the ones obtained by these other approaches:

\begin{itemize}
\item {\bf Static Word2Vec (SW2V)}~\cite{SGNS}: 
a static W2V is trained with the whole corpus (irrespective of the slices).
\item {\bf Transformed Word2Vec (TW2V)}~\cite{kulkarni2015statistically}, {\bf Aligned Word2Vec (AW2V)}~\cite{Hamilton2016} and {\bf Dynamic Word Embedding (DW2V)}~\cite{yao2018dynamic}: 
see previous work section for the details. 
\end{itemize}

As a \emph{baseline} for the equivalence metrics we consider the proportion of relations in the alignment test where $w = w'$ ($(s,w) \leftrightarrow (s',w)$), that is, those words for which its meaning did not change from significantly year $s$ to year $s'$.

\begin{table}[t]
\caption{Normalized Mutual Information for different embedding models. Results for SWV2, TW2V, AW2V, and DW2V models were obtained from~\cite{yao2018dynamic}.}\label{tab:NMI}
\centering
\begin{tabular}{@{}c|c|c|c|c@{}}
\toprule
\multirow{2}{*}{Dataset} & \multirow{2}{*}{Method} & \multicolumn{3}{c}{NMI} \\ \cmidrule(l){3-5} 
 &  & $k=10$  & $k=15$ & $k=20$  \\ \midrule
\multirow{5}{*}{NYT} & SW2V & 0.6736 & 0.6867 & 0.6713 \\
 & TW2V & 0.5175 & 0.5221 & 0.5130 \\
 & AW2V & 0.6580 & 0.6618 & 0.6386 \\
 & DW2V & 0.7175 & 0.7162 & 0.6906 \\
 & MW2V & \textbf{0.8840} & \textbf{0.8298} & \textbf{0.8045} \\ \midrule
TG & MW2V & 0.8661 & 0.8848 & 0.8406 \\ \midrule
NYT - TG & MW2V & 0.5866 & 0.6136 & 0.6113 \\ \bottomrule
\end{tabular}
\end{table}

\begin{table}[t]
\caption{$F_\beta$-score for different embedding models. Results for SWV2, TW2V, AW2V, and DW2V models were obtained in~\cite{yao2018dynamic}.}\label{tab:FB}
\centering
{\small
\begin{tabular}{@{}ccccc@{}}
\toprule
\multirow{2}{*}{Dataset} & \multirow{2}{*}{Method} & \multicolumn{3}{c}{$F_\beta$-score} \\ \cmidrule(l){3-5} 
 &  & \multicolumn{1}{c|}{$k=10$} & \multicolumn{1}{c|}{$k=15$} & \multicolumn{1}{c}{$k=20$} \\ \midrule
\multicolumn{1}{c|}{\multirow{5}{*}{NYT}} & \multicolumn{1}{c|}{SW2V} & \multicolumn{1}{c|}{0.6163} & \multicolumn{1}{c|}{0.7142} & 0.7214 \\
\multicolumn{1}{c|}{} & \multicolumn{1}{c|}{TW2V} & \multicolumn{1}{c|}{0.4584} & \multicolumn{1}{c|}{0.5072} & 0.5373 \\
\multicolumn{1}{c|}{} & \multicolumn{1}{c|}{AW2V} & \multicolumn{1}{c|}{0.6530} & \multicolumn{1}{c|}{0.7115} & 0.7187 \\
\multicolumn{1}{c|}{} & \multicolumn{1}{c|}{DW2V} & \multicolumn{1}{c|}{0.6949} & \multicolumn{1}{c|}{\textbf{0.7515}} & \textbf{0.7585} \\
\multicolumn{1}{c|}{} & \multicolumn{1}{c|}{MW2V} & \multicolumn{1}{c|}{\textbf{0.8190}} & \multicolumn{1}{c|}{0.6250} & 0.5420 \\ \midrule
\multicolumn{1}{c|}{TG} & \multicolumn{1}{c|}{MW2V} & \multicolumn{1}{c|}{0.9157} & \multicolumn{1}{c|}{0.8074} & 0.6470 \\ \midrule
\multicolumn{1}{c|}{NYT-TG} & \multicolumn{1}{c|}{MW2V} & \multicolumn{1}{c|}{0.6038} & \multicolumn{1}{c|}{0.4824} &  0.4231\\ \bottomrule
\end{tabular}
}
\end{table}

\begin{table}[th]
\caption{Similarity test for different embedding models. Results for SWV2, TW2V, AW2V, and DW2V models were obtained in~\cite{yao2018dynamic}.}\label{tab:sim}
\centering
{\footnotesize
\begin{tabular}{@{}c|c|c|c|c|l|l@{}}
\toprule
\multirow{2}{*}{Dataset} & \multirow{2}{*}{Method} & \multicolumn{5}{c|}{Similarity Tests} \\ \cmidrule(l){3-7} 
 &  & MP@1 & MP@3 & MP@5 & MP@10 & MRR \\ \midrule
\multirow{5}{*}{\begin{minipage}[c]{10mm}NYT tests\_1\end{minipage}} & SW2V & 0.2664 & 0.4210 & 0.4774 & 0.5612 & 0.3560 \\
 & TW2V & 0.0500 & 0.1168 & 0.1482 & 0.1910 & 0.0920 \\
 & AW2V & 0.1066 & 0.1814 & 0.2241 & 0.2953 & 0.1582\\
 & DW2V & 0.3306 & 0.4854 &0.5488 & 0.6191 & 0.4222\\
 & MW2V & \textbf{0.3929} & \textbf{0.5259} & \textbf{0.5767} & \textbf{0.6302} & \textbf{0.4702}\\ \midrule
\multirow{5}{*}{\begin{minipage}[c]{10mm}NYT tests\_2\end{minipage}} & SW2V & 0.0000 & 0.0787 & 0.0787 & 0.2022 & 0.0472 \\
 & TW2V & 0.0404 & 0.0764 & 0.0989 & 0.1438 & 0.0664 \\
 & AW2V & 0.0225 & 0.0517 & 0.0787 & 0.1416 & 0.0500 \\
 & DW2V & \textbf{0.0764} & 0.1596 & 0.2202 & \textbf{0.3820} & \textbf{0.1444} \\
 & MW2V & 0.0745 & \textbf{0.1677} & \textbf{0.2453} & 0.3385 & 0.1411 \\ \midrule
\multirow{2}{*}{TG} & Baseline & 0.4665 & - & - & \multicolumn{1}{c|}{-} & \multicolumn{1}{c}{-} \\
 & MW2V & 0.6093 & 0.7437 & 0.7750 & \multicolumn{1}{c|}{0.8169} & \multicolumn{1}{c}{0.6815} \\ 
\bottomrule
\end{tabular}
}
\end{table}

\begin{figure}[t]
    \centering
\includegraphics[width=0.5\columnwidth]{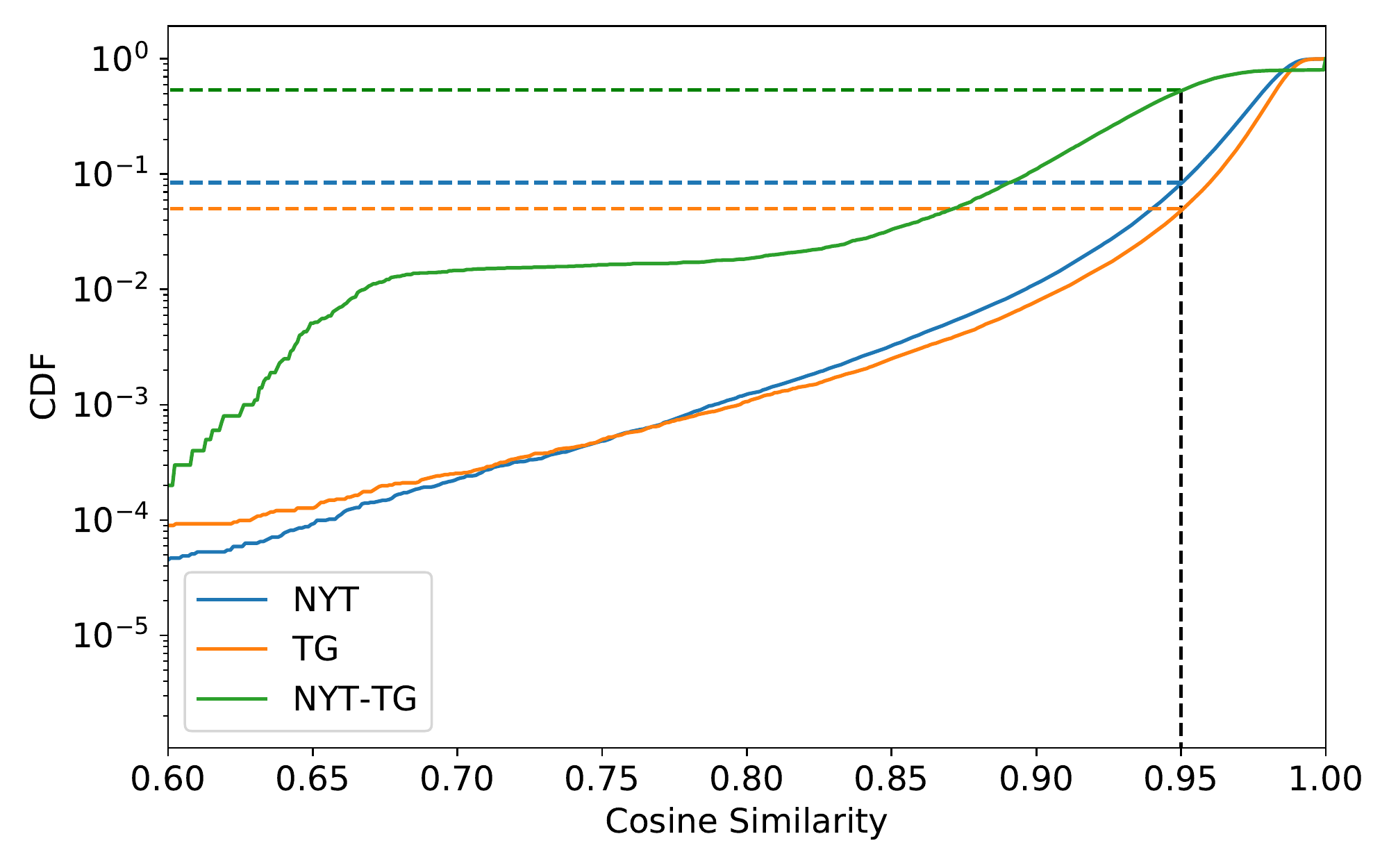}
    \caption{Cumulative distribution for the cosine similarities $\sigma(v_{s,w},\bar{v}_{w})$ (see Equation~\ref{eq:cos}), for the NTY, TG and NYT-TG datasets.}
    \label{fig:CDFs}
\end{figure}

\begin{figure}[t]
    \centering
\includegraphics[width=7.5cm]{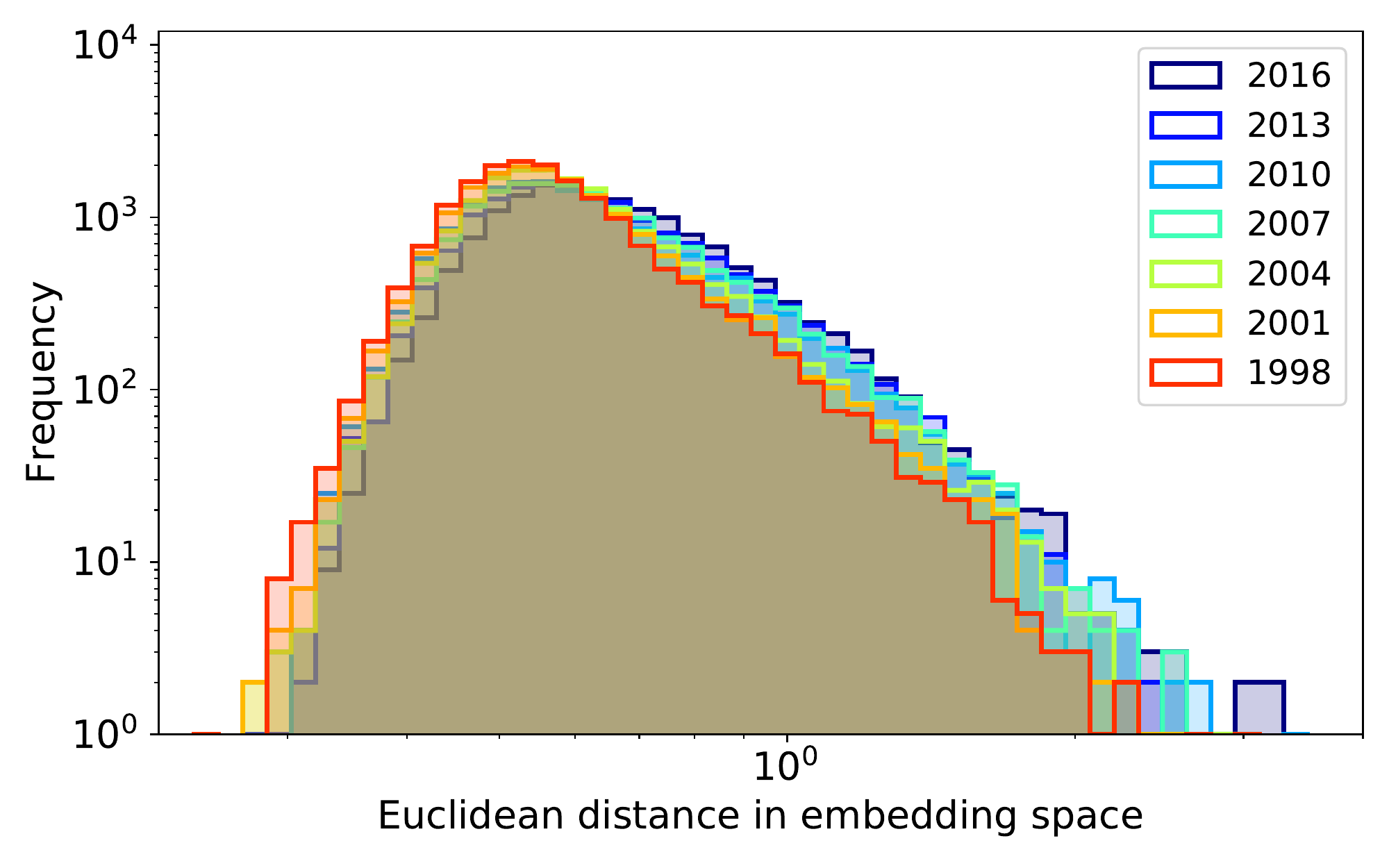}
\includegraphics[width=7.5cm]{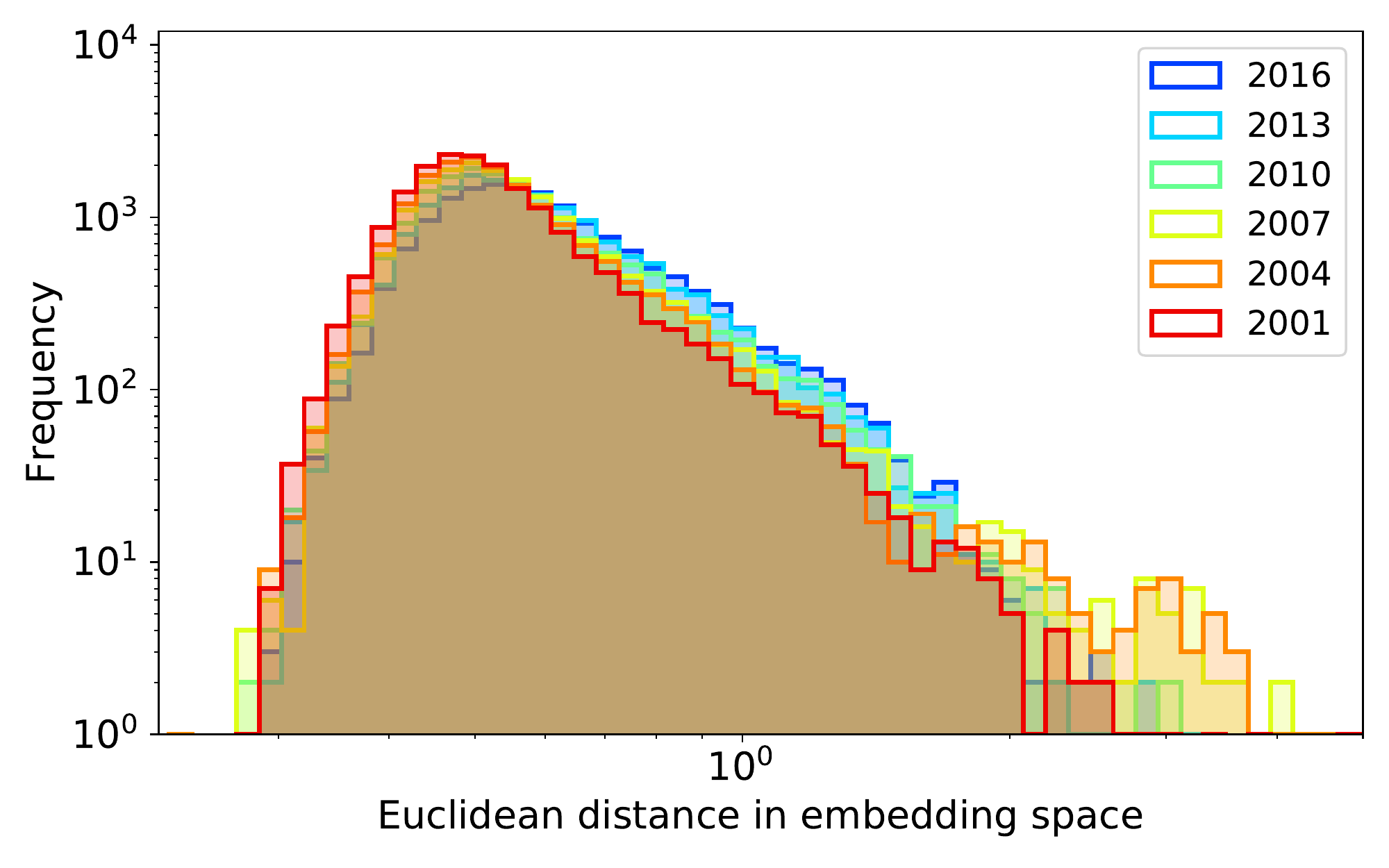}
    \caption{(top) Histogram of distances between word representations at $t_0=1995$ and different years for the New York Times dataset. (bottom) Same histogram for the dataset of The Guardian, with $t_0=1999$. }
    \label{fig:histograms}
\end{figure}

Table~\ref{tab:NMI} displays the $NMI$ results for the SW2V, TW2V, AW2V, DW2V, and our proposed MW2V. 
Notice that MW2V outperforms the others in the case of the NYT dataset, and the best score is for $k=10$ (we used $11$ sections). 
We also display our results for TG and NYT-TG datasets. 
Due to differences among datasets, the values are not comparable to those of the NYT. 
The TG case presents promising outcomes: its highest value is at $k=15$, close to the $17$ sections we considered (what is  consistent with the dataset).\\
The third dataset also got also its highest score for $k=15$, which is likewise close to the number of sections considered ($14$). 

Table~\ref{tab:FB} exhibits the $F_\beta$ score of the models. 
Notice MW2V's behavior is markedly better in the NYT dataset at $k=10$, while DW2V outperforms he other at $k=15$ and $k=20$.
In the TG case, MW2V's best values are obtained for $k=10$, the same as for the NYT-TG dataset. 
Following~\cite{yao2018dynamic} we chose $\beta=5$ in order to penalize false negatives, and probably the elements in that set could be higher for larger $k$.  

These two tests, $NMI$ and $F_\beta$, show that MW2V can achieve the semantic similarity across the time because of its ability to maintain close the vector representation of words along with the slices. 

Finally, Table~\ref{tab:sim} shows the similarity test. 
The MW2V presents the best results in {\em test\_1}, and similar outcomes to DW2V in {\em test\_2} for the NYT dataset.
These two methods, DW2V and MW2V, obtained ostensibly better scores than the other ones. 
In the case of the TG dataset, the outcome for MP\@1 is more than $30\%$ higher than the baseline.

In conclusion MW2V presented a performance as good as the one observed in \cite{yao2018dynamic} on alignment quality, being over it in some contexts.

\begin{figure}[t]
    \centering
    \hspace{-4mm}
\includegraphics[width=0.5\columnwidth,trim={5mm 10mm 10mm 21mm},clip]{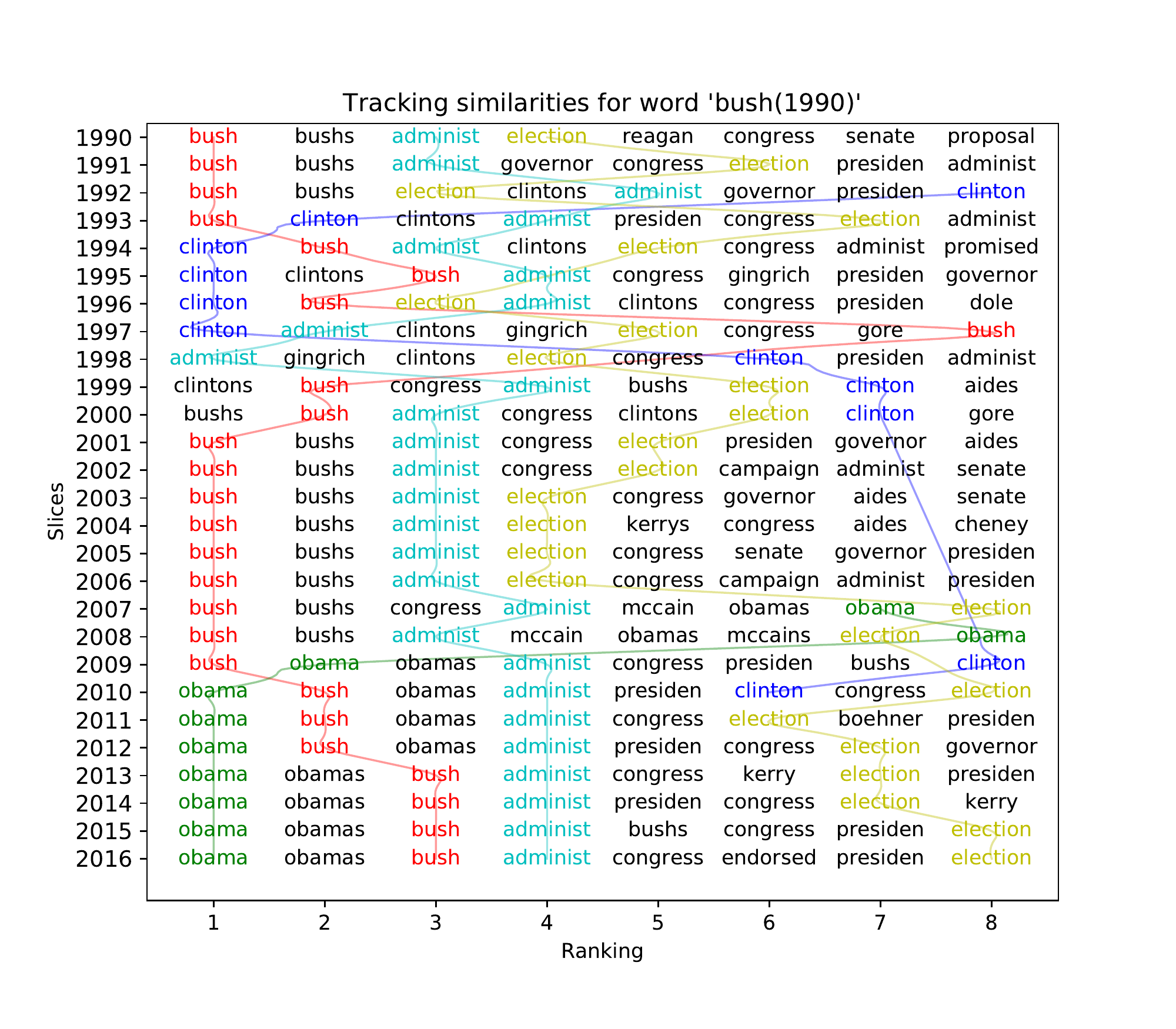}
    \caption{Semantic tracking over the years in NTY: the top-8 closest words to the representation of "Bush" in 1990. 
    Words are displayed with 8 characters, then ``adminst'' stands for a ``administration'' and the second occurrence for ``administrations''.
    Notice the first position is related to the USA's presidents.}
    \label{fig:NYT1}
\end{figure}

\begin{figure}[t]
    \centering
    \hspace{-4mm}
\includegraphics[width=0.5\columnwidth,trim={5mm 5mm 13mm 15mm},clip]{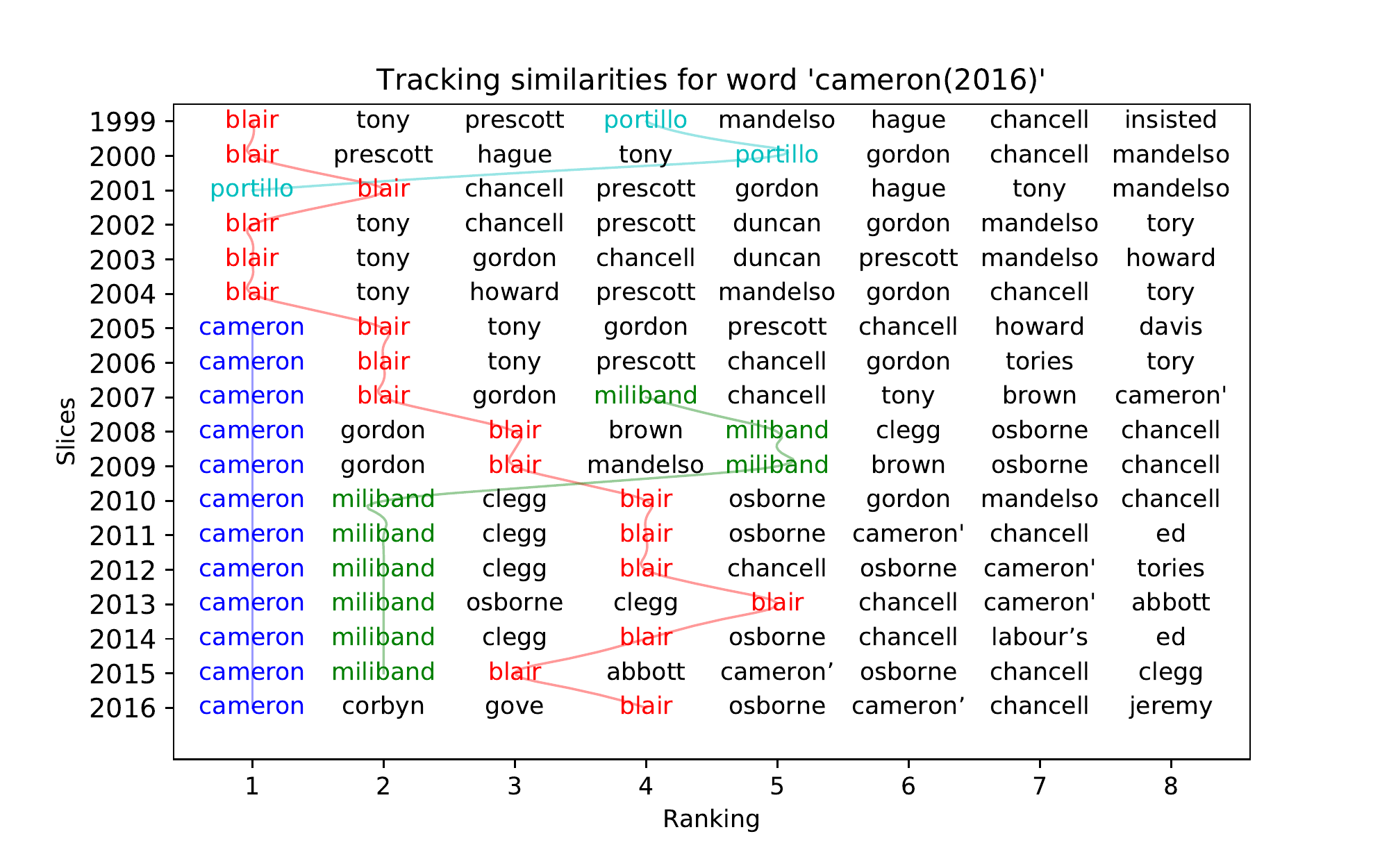}
    \caption{Semantic tracking over the years in TG: the top-8 closest words to the representation of ``Cameron'' in 2016. 
    Due to words are displayed with 8 characters, ``cameron'" stands for ``cameron's''.
 Notice heads positions are associated with politics.}
    \label{fig:TG1}
\end{figure}

\begin{figure}[t]
    \centering
    \hspace{-4mm}
\includegraphics[width=0.5\columnwidth,trim={5mm 10mm 10mm 21mm},clip]{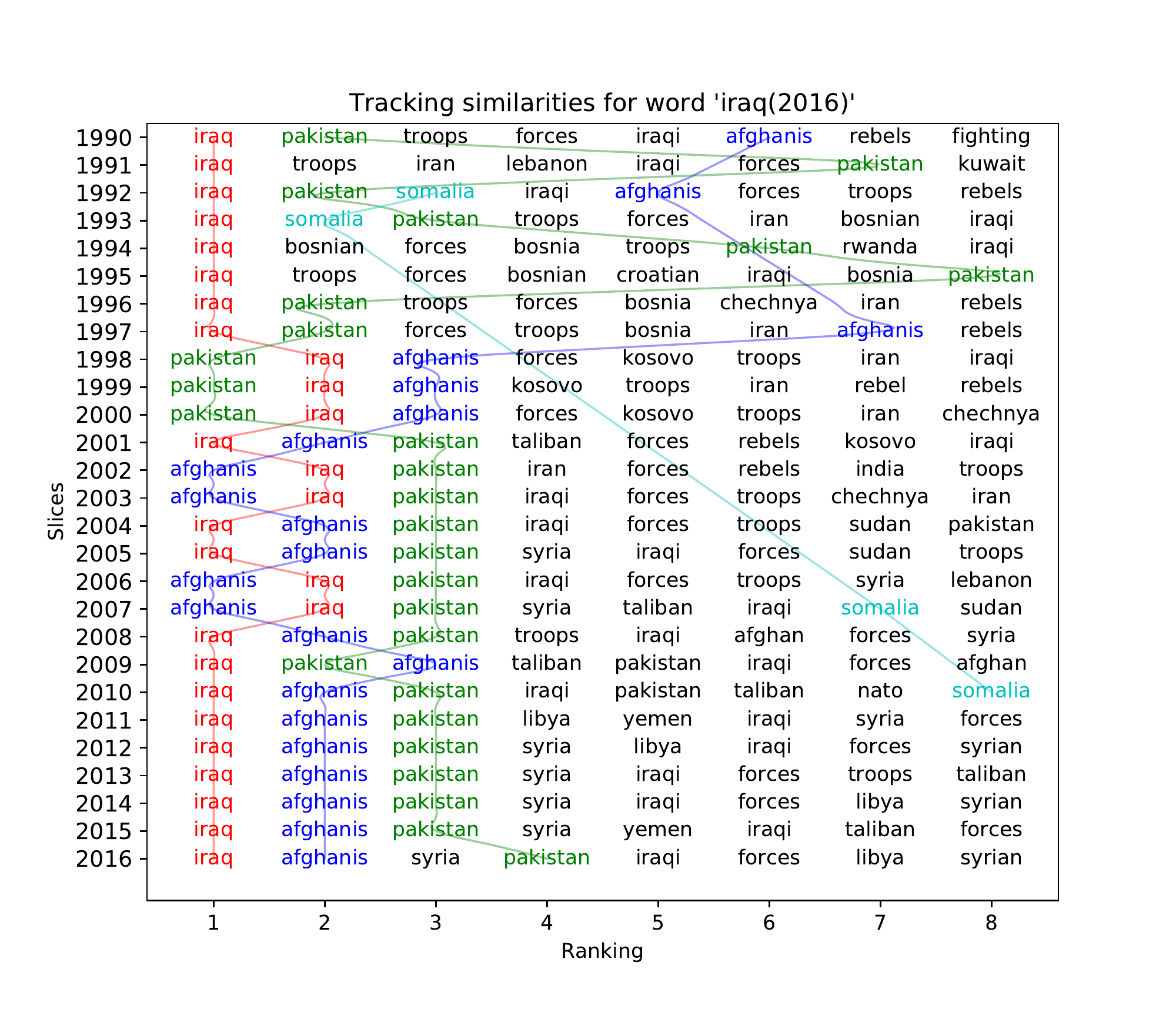}
    \caption{Semantic tracking over the years in NYT: the top-8 closest words to the representation of ``Iraq'' in 2016. 
    Words are limited to display the firs 8 characters, ``afghanis'' stands for ``afghanistan''.   
    Notice tops positions are correlated to military USA's conflicts.}
    \label{fig:NYT2}
\end{figure}

\begin{figure}[t]
    \centering
    \hspace{-4mm}
\includegraphics[width=0.5\columnwidth,trim={5mm 5mm 13mm 15mm},clip]{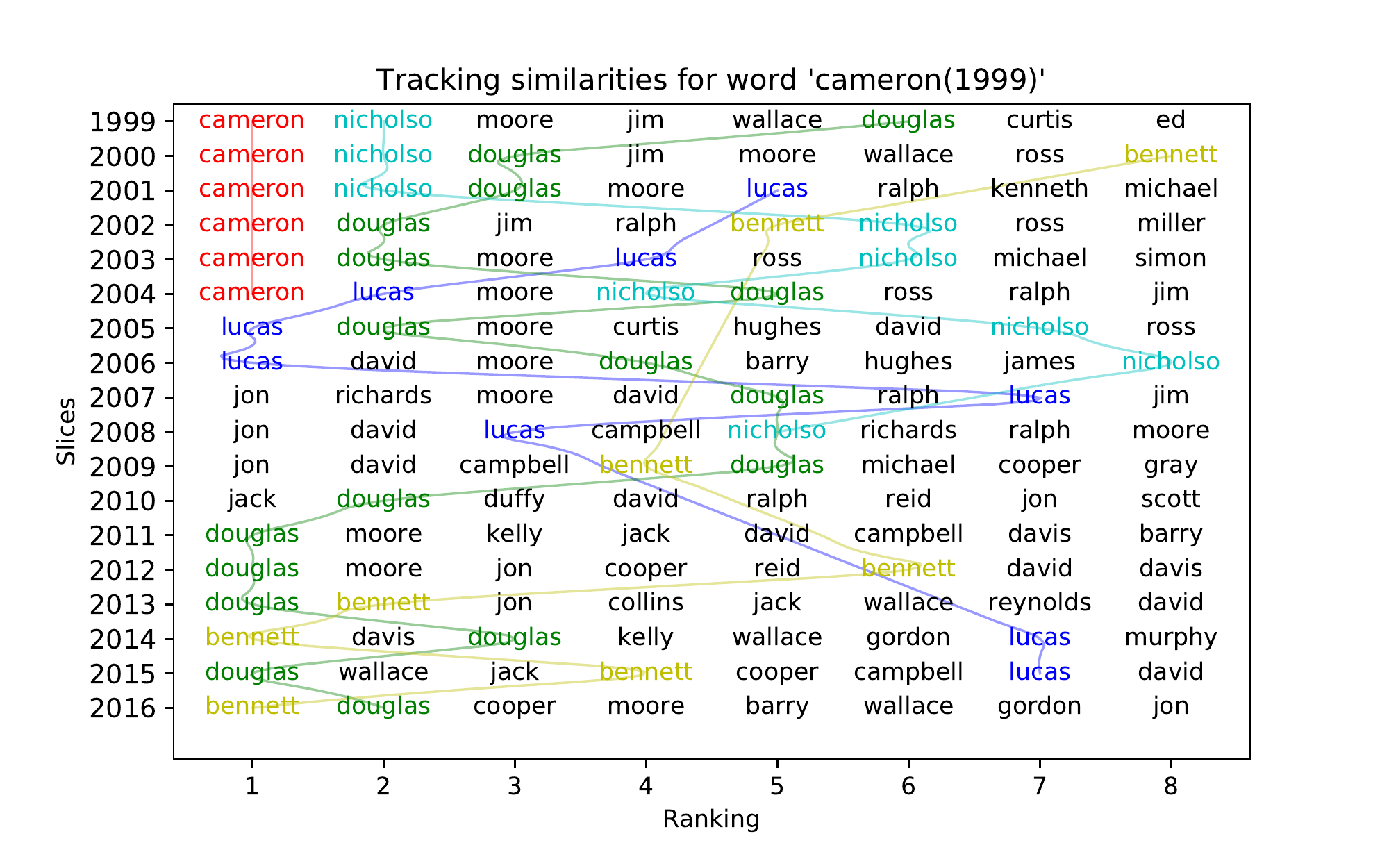}
    \caption{Semantic tracking over the years in TG: the top-8 closest words to the representation of ``Cameron'' in 1999.
    As displayed words are limited to 8 characters, ``nicholso'' stands for ``nicholson''.
 Notice heads positions are related to film production.}
    \label{fig:TG2}
\end{figure}

\begin{figure}[t]
    \centering
    \hspace{-4mm}
\includegraphics[width=0.5\columnwidth,trim={5mm 5mm 13mm 15mm},clip]{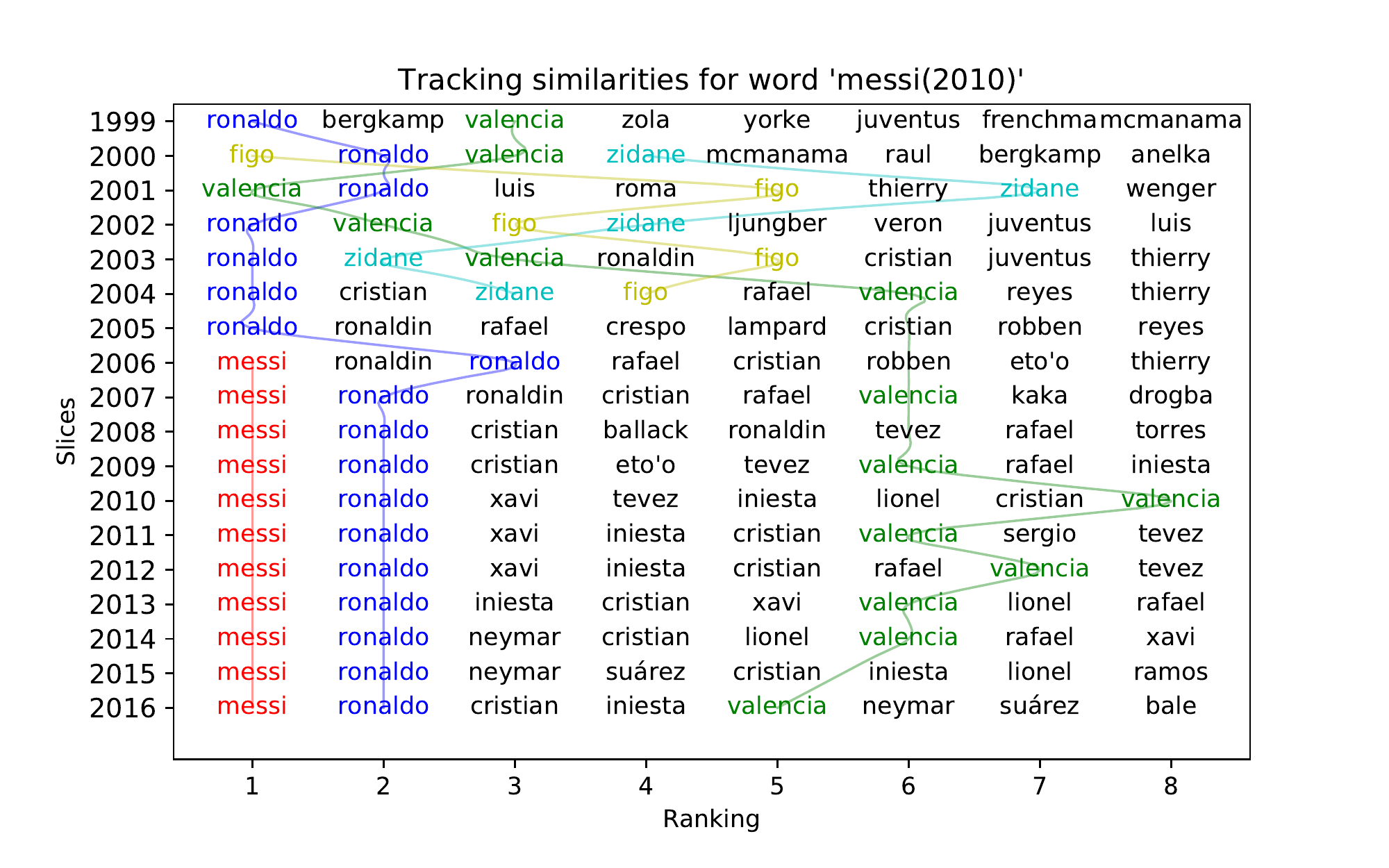}
    \caption{Semantic tracking over the years in TG: the top-8 closest words to the representation of ``Messi'' in 2010.
    The limitation to displaying 8 characters shows ``ronaldin'' instead ``ronaldinho''. 
 Notice leader positions are related to football.}
    \label{fig:TG3}
\end{figure}

\begin{figure}[t]
\includegraphics[width=0.5\columnwidth]{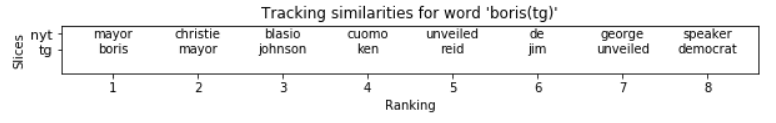}
\includegraphics[width=0.5\columnwidth]{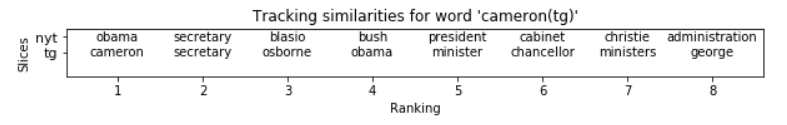}
\includegraphics[width=0.5\columnwidth]{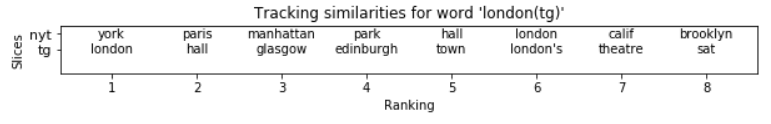}
\includegraphics[width=0.5\columnwidth]{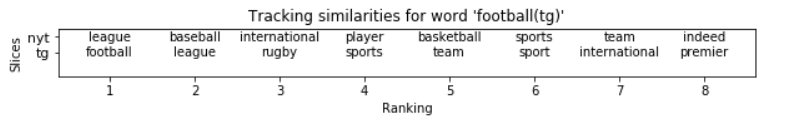}
\includegraphics[width=0.5\columnwidth]{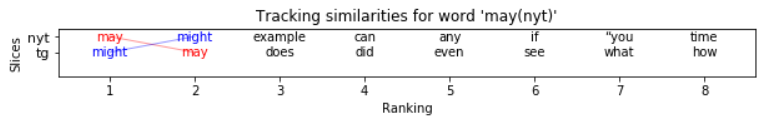}
\includegraphics[width=0.5\columnwidth]{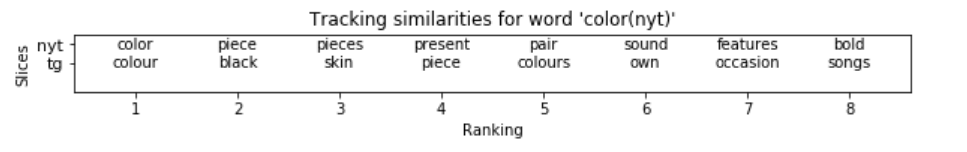}
\includegraphics[width=0.5\columnwidth]{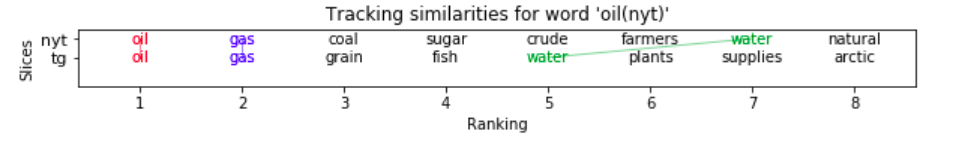}
\caption{Semantic tracking of the words in the NYT-TG corpus (2010-2016):  ``Boris'', ``Cameron'', ``London'', and ``football'' in TG; ``may'', ``color'', and ``oil'' in NYT. }
    \label{fig:NYT-TG}
    
\end{figure}

\subsection{Qualitative Evaluation}


We computed an average representation $\bar{v}_{w}$ for each word $w$, as the one minimizing the cosine similarity with the its slice representations $v_{s,w}$. This average representation can be obtained as:
$$\bar{v}_{w}=\sum_{s=1}^{S}\frac{v_{s,w}}{||v_{s,w}||} \enspace.$$
Note that the norm of this average representation is not relevant to us, but only its direction. Then, we compute the cosine similarity between each word representation $v_{s,w}$ and its average representation $\bar{v}_{w}$:
\begin{equation}
 \sigma(v_{s,w},\bar{v}_{w})= \frac{v_{s,w} \cdot \bar{v}_{w} }{ ||v_{s,w}|| \;  ||\bar{v}_{w}|| } \enspace. 
 \label{eq:cos}
\end{equation}
Figure~\ref{fig:CDFs} shows the distribution of these similarities $\sigma(v_{s,w},\bar{v}_{w})$ for every word appearing in at least eight slices. 
Most of the word representations have a cosine similarity greater than $0.95$, i.e., $91.6\%$ of the representations for the NYT, $95\%$ for TG, and $46.5\%$ for the NYT-TG case.
This implies that most of the words conserve their meaning through the slices, while only a relatively short proportion of them show a variation.


Then, we analyzed how well MW2V captures semantic change when applied to temporal slices. 
Following Bamler \& Mandt~\cite{bamler2017dynamic}, we computed histograms of the euclidean distance between words' representations in a base year and their representations some years later. Figure~\ref{fig:histograms} displays these histograms for the NYT and TG datasets. They show the that distance from the base year representation increases smoothly through the years, as expected.
This is non trivial, as our multi-source representation is not aware of the ordering in the temporal axis.

Our second analysis is centered on the search for equivalences.
Since embeddings capture relationships among words, most words should conserve their representation or either change it smoothly through the slices (e.g., who are the people in a certain political role, or the main sport figures). 
In each of the following comparisons we use the following methodology: we consider a particular representation of word $w$ in slice $s$, $v_{s,w}$ and then we look for the top-8 closest words to that representation in the other slices.
In the figures, we also highlight the trajectories of words in the ranking across the different slices.

Figures~\ref{fig:NYT1} and~\ref{fig:TG1} displays political personalities in US and UK respectively. 
The findings for the first position in the rank are similar to those in~\cite{yao2018dynamic}. 
We added words up to the eighth position in the neighborhood of each analyzed word.
In the NYT case, we found the elected president and his/her possessive name in the top rankings, some republican institutions, and other political personalities. 
Notice that it can detect both Bush's presidencies, father and son. 
The TG differs mainly because of newspaper public's orientation and British culture.
In this case, we can identify the prime minister's name is followed by other politicians' names, who eventually becomes the first minister. 

Regarding other aspects, Figure~\ref{fig:NYT2} presents some military conflicts in the NYT database.
We can follow the names of countries in conflict and see how their importance changes over time, and we also recognize military vocabulary.
Some instances have a smooth variation until becoming the top ones.

Then, Figure~\ref{fig:TG2} displays cinema directors, and Figure~\ref{fig:TG3} shows European football personalities in the TG dataset. 
The first figure shows how the search for a cinema director can obtain consistent results, even when having the same name as a political figure at a different time frame (i.e., Cameron in 2016 was the First Minister). 
Indeed, Cameron 1999 is the filmmaker, and the close names are also cinema-related people. 
Finally, the second figure shows that prominent sportspeople in football (a popular sport in England) can track the best figures in the discipline along years. 

The results of the NYT-TG are shown in Figure~\ref{fig:NYT-TG}. 
We display several word searches in both slices, the NYT and the TG (2010-2016).
The word ``boris'' in TG is related to the Mayor of London (Boris Johnson occupied this post from 2008 until 2016), which is followed by ``mayor, johnson''. 
Close to this position in the NYT, we find ``mayor'', ``christie'', ``blasio'', where the second was the  New Jersey Governor and the third the Mayor of New York.
Here, the equivalence between both newspapers is not precise, maybe because the importance of London in the U.K. could be comparable to that of a state in the U.S.
Searching in the neighborhood of ``cameron'' in TG, we found ``obama'' in the NYT, who had the analogous responsibilities in those years in the US.
The next row shows ``london'' and ``york'', related to equivalent cities from the NYT and the TG media perspectives.
Interestingly we also find the equivalence between ``might'' in British and ``may" in American English. 
And finally, the relation between ``football'' for the U.K. and ``baseball'' for the U.S. as the most popular sport is found in the last row.









\section{Conclusions}\label{sec:conc}

We proposed a multi-source embedding model, MW2V, aimed at dealing with general language variations. 
To demonstrate its feasibility, we applied the MW2V to three newspaper datasets: The New York Times and The Guardian to study temporal variations, and a combination of both datasets to model cultural variations. 
We performed an exhaustive evaluation of the method in text analysis tasks finding good quantitative and qualitative results compared to the state of the art, even for the temporal case, when the MW2V does not specifically model the time direction.

Future work includes the analysis of other applications, oriented to the exploitation of datasets, and also the possible implications of the use of a regularization parameter dependent on the slices and words, $\lambda_{s,w}$ instead of a constant one.
Moreover, some more insight is needed to answer open questions raised by~\cite{kutuzov-etal-2018-diachronic} for this proposal, namely, to try a broader scope of languages and to evaluate its robustness. 

\section*{Data Availability}
The embeddings for each training, the triplets for evaluation, the test sets for computing similarities and a brief code for the analysis of the embeddings are available at \url{https://github.com/CoNexDat/mw2v}.

\section*{Acknowledgments}
This work was partially financed by the OpLaDyn grant obtained in the 4th round of the Trans-Atlantic Platform: Digging into Data Challenge (HJ-253570 of IF-2017-14123506-APN-DNCEII\#MCT), and by UBACyT 2018 20020170100421BA.

\bibliographystyle{plain}
\bibliography{lw2v}

\end{document}